\documentclass[11pt,a4paper]{article}
\usepackage[hyperref]{acl2021}
\usepackage{times}
\usepackage{latexsym}

\usepackage{graphicx}
\usepackage{booktabs}
\usepackage{makecell}

\usepackage{enumitem}
\setitemize{noitemsep,topsep=0pt,parsep=0pt,partopsep=0pt}
\usepackage{microtype}

\usepackage{caption}
\usepackage{subcaption}

\usepackage{color, colortbl}
\definecolor{LightCyan}{rgb}{0.88,1,1}
\definecolor{LightRed}{rgb}{1.0, 0.91, 0.91}
\definecolor{LightGray}{rgb}{0.88,0.88,0.88}
\definecolor{VeryLightGray}{rgb}{0.93,0.93,0.93}

\aclfinalcopy % Uncomment this line for the final submission
\title{Unsupervised Pronoun Resolution via Masked Noun-Phrase Prediction}

\author{
Ming Shen\thanks{~~Equal Contribution} \quad Pratyay Banerjee\footnote[1]{}  \quad Chitta Baral  \\
Arizona State University \\
\texttt{mshen16, pbanerj6, chitta}@asu.edu               \\
}

\date{}

\begin{document}
\maketitle

\begin{abstract}
In this work, we propose Masked Noun-Phrase Prediction (MNPP), a pre-training strategy to tackle pronoun resolution in a fully unsupervised setting. Firstly, We evaluate our pre-trained model on various pronoun resolution datasets without any finetuning. Our method outperforms all previous unsupervised methods on all datasets by large margins. Secondly, we proceed to a few-shot setting where we finetune our pre-trained model on WinoGrande-S and XS separately. Our method outperforms RoBERTa-large baseline with large margins, meanwhile, achieving a higher AUC score after further finetuning on the remaining three official splits of WinoGrande.  
\end{abstract}

\section{Introduction}
\label{ssec:intro}
Co-reference resolution is an important NLP task that aims to find all expressions that refer to the same entity in a text.  The resolution of an ambiguous pronoun, known as pronoun resolution, is a longstanding challenge for the NLU community and an essential step for various high-level NLP tasks such as natural language inference \cite{bowman-etal-2015-large,williams-etal-2018-broad}, question answering \cite{rajpurkar-etal-2016-squad}, and relation extraction \cite{zhang-etal-2017-position}.

The most successful approach to  pronoun resolution is first fine-tuning a large pre-trained language model such as BERT \cite{devlin-etal-2019-bert} or RoBERTa \cite{liu2019roberta} on a human-labeled  pronoun resolution dataset such as Definite Pronoun Resolution Dataset (DPR) \cite{rahman-ng-2012-resolving} or WinoGrande (WG) \cite{sakaguchi2020winogrande}, and then either directly transferring to a smaller dataset such as Winograd Schema Challenge (WSC) \cite{levesque2012winograd} or Pronoun Disambiguation Problems (PDP) \cite{morgenstern2016planning} or further finetuning on a downstream dataset such as SuperGLUE-WSC \cite{wang2019superglue}. However, all the pipelines above can not avoid the phase of pre-training on a large human-labeled  pronoun resolution dataset.
% , which requires a huge amount of human effort to obtain, and this has been a practical problem for supervised PR.
Crowd-sourced ``unbiased" labels that do not introduce annotation-artifacts~\cite{gururangan-etal-2018-annotation} are shown to be costly and challenging to collect,  requiring a well-designed annotation interface and dedicated annotators.
To this end, we propose the unsupervised Masked Noun-Phrase Prediction task to pre-train a language model without any  pronoun resolution training signal and directly transfer the pre-trained model to downstream datasets such as WSC.\footnote{We refer to unsupervised or zero-shot transfer as without training on any  pronoun resolution dataset.} Two examples of WSC are listed in Table \ref{tab:wsc-exam}. Our work improves on all previous unsupervised methods by large margins and even outperforms several strong supervised methods on all datasets we study.

\begin{table}[t]
\centering
\small
\resizebox{\linewidth}{!}{
\begin{tabular}{@{}ll@{}}
\toprule
\textbf{WSC Sentences} & \textbf{Candidate Choices} \\ \midrule
\begin{tabular}[c]{@{}l@{}} The trophy doesn't fit in the \\ suitcase because \textbf{it} is too \underline{small}.
\end{tabular} & \textbf{A.} the trophy \textbf{B.} the suitcase \\ \midrule 
\begin{tabular}[c]{@{}l@{}} The trophy doesn't fit in the \\ suitcase because \textbf{it} is too \underline{big}.
\end{tabular} & \textbf{A.} the trophy \textbf{B.} the suitcase \\ \bottomrule
\end{tabular}}
\caption{\small Above are two WSC examples. A system is required to resolve the bold pronoun \textbf{``it''} to ``the suitcase'' in the first sentence and to ``the trophy'' in the second sentence.}
\label{tab:wsc-exam}
\end{table}

We then proceed to the few-shot setting where we finetune our best zero-shot model on WinoGrande-S and XS respectively. MNPP gives a large margin of improvements over strong baselines including CSS \cite{klein-nabi-2020-contrastive}, RoBERTa-large \cite{sakaguchi2020winogrande}, and UnifiedQA-BART-large \cite{khashabi-etal-2020-unifiedqa}. We further finetune on the remaining three data splits and achieve a higher AUC score on all five splits of WinoGrande over RoBERTa-large baseline.

In summary, our main contributions in this work are threefold.

\begin{itemize} 
    \item \textbf{First}, we propose the MNPP pre-training task and study how different synthetic dataset properties affect zero-shot performances.
    \item  \textbf{Second}, we show MNPP outperforms all previous fully unsupervised methods and even several strong supervised baselines on all pronoun resolution datasets we study.
    \item  \textbf{Finally}, we show that under few-shot settings, MNPP pre-training gives a significant performance boost on WinoGrande-S and XS and furthermore achieves a higher AUC score over all five splits of WinoGrande.
\end{itemize}

\section{Related Works}
\label{ssec:relatedwork}
In this work, we mainly compare with unsupervised methods.\footnote{Please refer to supplemental materials for more details on supervised methods.}
On WSC, \citet{zhang2018distributed} propose the first unsupervised model where they modify Skip-Gram \cite{mikolov2013efficient} objective to predict semantic dependencies then use this additional information during testing. \citet{wang-etal-2019-unsupervised} propose Unsupervised Deep Structured Semantic Models (UDSSM), which utilizes BiLSTM \cite{hochreiter1997long} to compute contextual word embedding and uses models ensemble. \citet{klein-nabi-2019-attention} directly explore the inner attention layers of BERT. \citet{ye2019align} adapt a masking and predicting strategy, called align, mask, and select (AMS), where entities that are connected with ConceptNet \cite{speer2012representing} are masked and the model is required to select from a given list of candidate entities. An ensemble of large pre-trained models is first utilized by \citet{trinh2018simple}. GPT-2 is directly evaluated on WSC in \citet{radford2019language}. \citet{prakash-etal-2019-combining} extend a language model with a knowledge hunting strategy. \citet{kocijan-etal-2019-surprisingly} and \citet{kocijan-etal-2019-wikicrem} are the most similar works to us and we will discuss the details in Section \ref{ssec:mnppdiscussion}. Most recently, \citet{klein-nabi-2020-contrastive} study a contrastive self-supervised learning approach (CSS) for WSC and DPR and also establish the first unsupervised baseline for KnowRef~\cite{emami-etal-2019-knowref}. On WinoGrande, knowledge hunting \cite{prakash-etal-2019-combining} and language models ensemble \cite{sakaguchi2020winogrande} have been studied.

\begin{table*}[t]
\centering
\small
% \resizebox{\linewidth}{!}{
\begin{tabular}{@{}lccccc@{}}
\toprule
\textbf{Dataset $\backslash$ Source} & \textbf{CNN} & \textbf{QUOREF} & \textbf{Gutenberg} & \textbf{Knowledge} & \textbf{Total} \\ \midrule
Hybrid Source & 100,556 & 51,451 & 6,381 & - & 158,388 \\
\begin{tabular}{@{}l@{}}Hybrid Source  \\ w/ Knowledge   \end{tabular} & 189,376 & 98,844 & 19,424 & 75,993 & 383,637 \\ \bottomrule                        
\end{tabular}
% } 
\caption{\small Number of instances from each source of two hybrid-source synthetic datasets in the first group.}
\label{tab:synstat}
\end{table*}

\begin{table*}[th!]
\small
\centering
\scalebox{1}{
\begin{tabular}{@{}lccccc@{}}
\toprule
\textbf{Synth. Dataset $\backslash$ Downstream} & \textbf{WinoGrande (AUC)} & \textbf{WSC} & \textbf{DPR} & \textbf{KnowRef} & \textbf{COPA} \\ \midrule
Hybrid Source (160k) & 58.08 (\textbf{0.6961}) & \textbf{79.48} & 82.27 & 79.83 & 71.29 \\
\makecell{Hybrid Source w/ Know. (380k)} & 58.56 (0.6821) & 78.39 & \textbf{83.88} & 79.04 & 73.27 \\ \midrule
Gutenberg-10k & 57.93 (-) & 75.09 & 81.21 & 77.15 & 79.21 \\ 
Gutenberg-50k & 57.40 (-) & 76.19 & 77.84 & 75.10 & 74.26 \\ 
Gutenberg-100k & 58.56 (-) & 72.53 & 75.00 & 74.40 & 75.25 \\ 
Gutenberg-300k & 57.38 (-) & 75.82 & 81.56 & 76.44 & 78.22 \\ 
Gutenberg-500k & \textbf{59.19} (0.6748) & 76.56 & 80.50 & 79.12 & \textbf{85.51} \\ \midrule
Gutenberg-Easy (33k) & 56.43 (-) & 69.60 & 70.92 & 75.10 & 77.23 \\ 
Gutenberg-Medium (33k) & 57.00 (-) & 75.10 & 80.32 & 78.17 & 79.21 \\ 
Gutenberg-Hard (33k) & 57.54 (-) & 75.82 & 80.67 & \textbf{79.98} & 74.36 \\ \bottomrule 
\end{tabular}} 
\caption{\small Zero-shot transfer performances (\%) on downstream datasets. AUC scores of WinoGrande are calculated after finetuning on all 5 splits of WinoGrande training sets. Difficulty level is decided using cosine similarity between the two candidate word vectors. Hard samples are the top 33\% of samples when they are sorted in descending order using similarity score. Easy are bottom 33\%, with Medium in-between.}
\label{tab:unsup}
\end{table*}

\section{Masked Noun-Phrase Prediction}
\label{ssec:mnpp}
We treat MNPP as a binary classification task. Given the sentence: \textit{``She put the cup on the chair, but he knocked over \underline{the chair}, and the cup fell.''}, the underlined \textit{``\underline{the chair}''} will be masked and a pair of replacement phrases for this masked position is given as \textit{\{``the cup'', ``the chair''\}}. One of the candidates is the masked phrase,\textit{``the chair''}, and the other candidate is a different phrase in the sentence, \textit{``the cup''} extracted from \textit{``She put the cup on the chair''}. The constraint we impose is that both the ground-truth noun-phrase and the alternative candidate need to appear before the masked phrase location, which mimics the pronoun resolution task. We sample sentences following the above constraint to create our synthetic datasets for pre-training. 

We convert the sentence into the format of \{[CLS] \textit{first\_half} \textbf{option} \textit{second\_half} [SEP]\} where \textit{first\_half} refers to \textit{``She put the cup on the chair but he knocked over ''} and \textit{second\_half}  refers to \textit{``, and the cup fell.''}. The \textbf{option} is replaced by candidates, \textit{``the cup''} or \textit{``the chair''}. We compute P(\textit{the chair}$|$\textit{sentence}, $\theta$) and P(\textit{the cup}$|$\textit{sentence}, $\theta$) and optimize $\theta$, the parameters of the model, using cross-entropy loss. We use the final layer [CLS] vector from transformer-based language models and pass it through a single layer feed-forward network to calculate the logits.

% \noindent
% \textbf{Discussion:} 
\subsection{Discussion}
\label{ssec:mnppdiscussion}
The intuition behind MNPP is that given sufficient samples that mimic pronoun resolution task, the model can learn rich knowledge to perform well on human-annotated pronoun resolution datasets. Such idea is also in-line with recent advances in unsupervised QA \cite{lewis-etal-2019-unsupervised,li-etal-2020-harvesting,banerjee-baral-2020-self,banerjee2020self, banerjee-etal-2021-self}, where synthetic QA datasets are created from unannotated corpora to perform unsupervised pre-training. 
Strictly speaking, MNPP is even more unsupervised since our synthetic datasets are not created with true pronoun resolution signals, whereas synthetic QA datasets in works cited above contain true question-answer pairs.

As mentioned in previous Section \ref{ssec:relatedwork}, similar to our work, \citet{kocijan-etal-2019-surprisingly} studied such pre-training strategy by constructing a synthetic dataset, called MaskedWiki, which is crawled from English Wikipedia. However, our work is significantly different from theirs in the following ways. First, their pipeline requires further finetuning on another pronoun resolution task before transferring to downstream datasets, whereas our method can be directly evaluated on downstream datasets. Second, the size of MaskedWiki is 2.4 millions, which is 15 times the size of our best performing synthetic dataset. Third, we study how different properties of synthetic datasets affect zero-shot performances. Finally, they use a masked token prediction loss, and we model it as a classification task. \citet{kocijan-etal-2019-wikicrem} also construct another synthetic dataset called WikiCREM following the same masking principle but with only personal names masked.

% There is also a slight similarity between MNPP pre-training and the classification task of the discriminator in ELECTRA \cite{Clark2020ELECTRA:}. The difference is that the phrases we mask are constrained to be noun-phrases since our focus is pronoun resolution, however, tokens are randomly masked by the generator in ELECTRA.

\section{Experiments and Results}

% \noindent
% \textbf{Synthetic Dataset:}
\subsection{Synthetic Dataset}
\label{ssec:syntheticdatset}
We study three properties of synthetic dataset: source style, size, and difficulty level. The sources we choose include various styles of texts, including CNN stories \cite{see-etal-2017-get}, Wikipedia, and PG-19 language modeling benchmark \cite{Rae2020Compressive}. We study 3 groups and a total of 10 different synthetic datasets. The first group contains two synthetic datasets collected from all sources with and without knowledge hunting strategy \cite{prakash-etal-2019-combining}. The second group contains five synthetic datasets collected only from PG-19 but with varying sizes from 10k to 500k. The third group contains three synthetic datasets collected from PG-19 but with easy, medium, and hard samples with the same size of 33k each.\footnote{Please refer to supplemental materials for details on synthetic datasets constructions.} Datasets' names are listed in the first column of Table \ref{tab:unsup} and statistics of the first group are described in Table \ref{tab:synstat}.

% Details of data construction are described in Appx. \ref{ssec:appxsynthdataset}.

% \noindent
% \textbf{Unsupervised Pronoun Resolution:} 
\subsection{Unsupervised Pronoun Resolution}
The downstream datasets we test on are the WinoGrande test set (17k instances), DPR test set (564 instances), KnowRef test set (12k instances), and COPA validation set (101 instances). Although COPA \cite{wang2019superglue} is a cause and effect identification dataset, \citet{sakaguchi2020winogrande} show that directly transferring from a WinoGrande-finetuned RoBERTa-large model to COPA already achieves a good performance, indicating that finetuning on WinoGrande can serve as a resource for common sense knowledge. We also investigate whether learning through MNPP can serve as a resource for common sense. Note that we also provide evaluation on the GAP dataset \cite{webster-etal-2018-mind} in Table \ref{tab:gap} for reference although the authors of GAP explicitly mention in their paper that they urge the community to not treat GAP as a Winograd-style task but a co-reference resolution task without gold mention provided. 
% Experiment details are included in Appx. %\ref{ssec:zeroexpdetails}.

\begin{table}[t]
\small
\centering
\begin{tabular}{@{}lc@{}}
\toprule
\multicolumn{2}{c}{\textbf{WSC} \cite{levesque2012winograd}}\\ \midrule
\rowcolor{VeryLightGray} \underline{Bi-LSTM-DPR \shortcite{opitz-frank-2018-addressing}} & 56.0 \\
\rowcolor{VeryLightGray} \underline{BERT\_NSP-DPR \shortcite{ruan2019exploring}} & 71.1 \\
\rowcolor{VeryLightGray} \underline{CorefBERT\textsubscript{LARGE} \shortcite{ye-etal-2020-coreferential}} & 71.4 \\
\rowcolor{VeryLightGray} \underline{BERT-WIKICREM-DPR \shortcite{kocijan-etal-2019-wikicrem}} & 71.8 \\
\rowcolor{VeryLightGray} \underline{BERT-MASKEDWIKI-DPR \shortcite{kocijan-etal-2019-surprisingly}} & 72.5 \\
\rowcolor{VeryLightGray} \underline{UDSSM-MASKEDWIKI-DPR \shortcite{he2019hybrid}} & 75.1 \\
\rowcolor{VeryLightGray} \underline{AMS-CSQA-DPR \shortcite{ye2019align}} & 75.5 \\
\rowcolor{VeryLightGray} RoBERTa-DPR \shortcite{sakaguchi2020winogrande} & 83.1 \\
\rowcolor{VeryLightGray} CorefRoBERTa\textsubscript{LARGE} \cite{ye-etal-2020-coreferential} & 83.2 \\
\rowcolor{VeryLightGray} RoBERTa-WG \shortcite{sakaguchi2020winogrande} & \textbf{90.1} \\ %\midrule
\rowcolor{LightCyan} Modified Skip-Gram \shortcite{zhang2018distributed} & 60.3 \\
\rowcolor{LightCyan} BERT Inner Attention \shortcite{klein-nabi-2019-attention} & 60.3 \\
\rowcolor{LightCyan} BERT-MASKEDWIKI \shortcite{kocijan-etal-2019-surprisingly} & 61.9 \\
\rowcolor{LightCyan} UDSSM \shortcite{wang-etal-2019-unsupervised} & 62.4 \\
\rowcolor{LightCyan} BERT-WIKICREAM \shortcite{kocijan-etal-2019-wikicrem} & 63.4 \\
\rowcolor{LightCyan} Ensemble LMs \shortcite{trinh2018simple} & 63.7 \\
\rowcolor{LightCyan} CSS \shortcite{klein-nabi-2020-contrastive} & 69.6 \\
\rowcolor{LightCyan} GPT-2 \shortcite{radford2019language} & 70.7 \\
\rowcolor{LightCyan} WSC Know. Hunting \shortcite{prakash-etal-2019-combining} & 71.1 \\
\rowcolor{LightCyan} \textbf{MNPP (this work)} & \textbf{79.5} \\ \bottomrule
\end{tabular}
\end{table}

\begin{table}[t]
\small
\centering
% \resizebox{\linewidth}{!}{
\begin{tabular}{@{}lcc@{}}
\toprule
\multicolumn{2}{c}{\textbf{WinoGrande} \cite{sakaguchi2020winogrande}} & \textbf{AUC} \\ \midrule
\rowcolor{VeryLightGray} \underline{RoBERTa (local context) \shortcite{sakaguchi2020winogrande}} & 50.0 & - \\
\rowcolor{VeryLightGray} \underline{BERT-DPR \shortcite{sakaguchi2020winogrande}} & 51.0 & - \\
\rowcolor{VeryLightGray} \underline{BERT (local context) \shortcite{sakaguchi2020winogrande}} & 51.9 & - \\
\rowcolor{VeryLightGray} \underline{RoBERTa-DPR \shortcite{sakaguchi2020winogrande}} & 58.9 & - \\
\rowcolor{VeryLightGray} BERT \shortcite{sakaguchi2020winogrande} & 64.9 & \underline{0.5289} \\
\rowcolor{VeryLightGray} CSS \shortcite{klein-nabi-2020-contrastive} & 65.0 & \underline{0.6046} \\
\rowcolor{VeryLightGray} UnifiedQA-Bart-large \shortcite{khashabi-etal-2020-unifiedqa} & 73.3 & \underline{0.6358} \\
\rowcolor{VeryLightGray} CorefRoBERTa\textsubscript{LARGE} \shortcite{ye-etal-2020-coreferential} & 77.9 & - \\
\rowcolor{VeryLightGray} RoBERTa-large \shortcite{sakaguchi2020winogrande} & 79.1 & \underline{0.6641} \\
\rowcolor{VeryLightGray} CorefBERT\textsubscript{LARGE} \shortcite{ye-etal-2020-coreferential} & 80.8 & - \\
\rowcolor{VeryLightGray} TTTTT \shortcite{lin2020tttttackling} & 84.6 & 0.7673 \\
\rowcolor{VeryLightGray} UnifiedQA-T5-11B \shortcite{khashabi-etal-2020-unifiedqa} & \textbf{89.4} & 0.8571 \\
\rowcolor{LightCyan} Wino Know. Hunting \shortcite{sakaguchi2020winogrande} & 49.6 & - \\
\rowcolor{LightCyan} Ensemble LMs \shortcite{sakaguchi2020winogrande} & 50.9 & - \\
\rowcolor{LightCyan} \textbf{MNPP (this work)} & \textbf{59.2} & 0.6706 \\ \bottomrule
\end{tabular}
% }
\end{table}

\begin{table}[t]
\small
\centering
% \resizebox{\linewidth}{!}{
\begin{tabular}{@{}lc@{}}
\toprule
\multicolumn{2}{c}{\textbf{DPR} \cite{rahman-ng-2012-resolving}} \\ \midrule
\rowcolor{VeryLightGray} \underline{Bi-LSTM \shortcite{opitz-frank-2018-addressing}} & 63.0 \\
\rowcolor{VeryLightGray} \underline{FeatureEng+Ranking \shortcite{rahman-ng-2012-resolving}} & 73.0 \\
\rowcolor{VeryLightGray} \underline{BERT-WIKICREM-DPR \shortcite{kocijan-etal-2019-wikicrem}} & 80.0 \\
\rowcolor{VeryLightGray} \underline{BERT-DPR \shortcite{kocijan-etal-2019-wikicrem}} & 83.3 \\
\rowcolor{VeryLightGray} BERT-MASKEDWIKI-DPR \shortcite{kocijan-etal-2019-surprisingly} & 84.8 \\
\rowcolor{VeryLightGray} BERT-WG \shortcite{sakaguchi2020winogrande} & 84.9 \\
\rowcolor{VeryLightGray} CorefBERT\textsubscript{LARGE} \cite{ye-etal-2020-coreferential} & 85.1 \\
\rowcolor{VeryLightGray} RoBERTa-DPR \shortcite{sakaguchi2020winogrande} & 91.7 \\
\rowcolor{VeryLightGray} CorefRoBERTa\textsubscript{LARGE} \cite{ye-etal-2020-coreferential} & 92.2 \\
\rowcolor{VeryLightGray} RoBERTa-WG \shortcite{sakaguchi2020winogrande} & 92.5 \\
\rowcolor{VeryLightGray} RoBERTa-WG-DPR \shortcite{sakaguchi2020winogrande} & \textbf{93.1} \\
\rowcolor{LightCyan} BERT-WIKICREAM \shortcite{kocijan-etal-2019-wikicrem} & 67.4 \\
\rowcolor{LightCyan} CSS \shortcite{klein-nabi-2020-contrastive} & 80.1 \\
\rowcolor{LightCyan} \textbf{MNPP (this work)} & \textbf{83.9} \\ \midrule
\multicolumn{2}{c}{\textbf{KnowRef} \cite{emami-etal-2019-knowref}} \\ \midrule
\rowcolor{VeryLightGray} \underline{E2E-CoNLL} \shortcite{emami-etal-2019-knowref} & 60.0 \\
\rowcolor{VeryLightGray} \underline{E2E-KnowRef} \shortcite{emami-etal-2019-knowref} & 61.0 \\
\rowcolor{VeryLightGray} \underline{BERT \shortcite{emami-etal-2019-knowref}} & 65.0 \\
\rowcolor{VeryLightGray} \underline{E2E-KnowRef+CoNLL \shortcite{emami-etal-2019-knowref}} & 65.0 \\
\rowcolor{VeryLightGray} RoBERTa-DPR \shortcite{sakaguchi2020winogrande} & 84.2 \\
\rowcolor{VeryLightGray} RoBERTa-WG \shortcite{sakaguchi2020winogrande} & \textbf{85.6} \\ 
\rowcolor{LightCyan} CSS \shortcite{klein-nabi-2020-contrastive} & 65.5 \\
\rowcolor{LightCyan} \textbf{MNPP (this work)} & \textbf{80.0} \\ \midrule
\multicolumn{2}{c}{\textbf{COPA} \cite{wang2019superglue}} \\ \midrule
\rowcolor{VeryLightGray} \underline{RoBERTa-WG \shortcite{sakaguchi2020winogrande}} & 84.4 \\
\rowcolor{LightCyan} \textbf{MNPP (this work)} & \textbf{85.5} \\ \bottomrule
\end{tabular}
% }
\caption{\small Comparisons of zero-shot transfer performance (\%) among baselines and MNPP. Works highlighted with gray are supervised methods either directly finetuned on downstream datasets or additionally finetuned on another pronoun resolution dataset. Works highlighted with cyan are fully unsupervised methods. Best performances are in bold. We also underline supervised methods that our method outperforms. Note that AUC score for MNPP is obtained after finetuning on all WinoGrande data splits. (Model-A-B stands for model finetuned on A and B sequentially.)}
\label{tab:unsupcompare}
\end{table}

% \smallskip
% \noindent
% \textbf{Results \& Discussion.}
% \noindent
% \textbf{Results:} 
\subsubsection{Results}
We report our experiment results in Table \ref{tab:unsup} and Table \ref{tab:unsupcompare}.
% The best zero-shot performance of each synthetic dataset on each downstream dataset is reported in Table \ref{tab:unsup}, and comparisons with previous works are listed in Table \ref{tab:unsupcompare}. 
Table \ref{tab:unsup} shows that different downstream dataset benefits from different property of the synthetic dataset.
The hybrid-source synthetic dataset of size 160k outperforms PG-500k by a large margin on both WSC and DPR. It shows that pre-training on text of various styles instead of larger size is probably a better guarantee for better zero-shot performance on WSC and DPR. 
However, on WinoGrande and KnowRef, text style and dataset size both seem to impact zero-shot performance. On WinoGrande, larger size matters slightly more, whereas on KnowRef, synthetic dataset with various styles of texts gives better performance. On COPA, it is clear that using books as the source and with larger size at the same time is the key, probably because fictional event descriptions describing day-to-day activities in books contain more common sense, whereas CNN or Wikipedia articles contain precise, factual, non-fictional event descriptions. Finally, pre-training on more challenging examples helps on all tasks except COPA.

\begin{table}[t]
\small
\centering
\resizebox{\linewidth}{!}{
\begin{tabular}{@{}lcccc@{}}
\toprule
    % \multicolumn{5}{c}{\textbf{GAP} \cite{webster-etal-2018-mind}}  \\ \midrule 
 & M & F & B & O \\ \midrule
\rowcolor{VeryLightGray} \begin{tabular}{@{}l@{}}BERT  \\ \cite{kocijan-etal-2019-wikicrem}\end{tabular} & 75.3 & 75.1 & 1.00 & 75.2 \\
\rowcolor{VeryLightGray} \begin{tabular}{@{}l@{}}CorefBERT\textsubscript{LARGE} \\ \cite{ye-etal-2020-coreferential}\end{tabular}   & - & - & - & 76.8 \\
\rowcolor{VeryLightGray} \begin{tabular}{@{}l@{}}BERT-WIKICREM-GAP  \\ \cite{kocijan-etal-2019-wikicrem}\end{tabular}  & 76.4 & 78.4 & 1.03 & 77.4 \\
\rowcolor{VeryLightGray} \begin{tabular}{@{}l@{}}CorefRoBERTa\textsubscript{LARGE}  \\ \cite{ye-etal-2020-coreferential}\end{tabular}   & - & - & - & 77.8 \\
\rowcolor{VeryLightGray} \begin{tabular}{@{}l@{}}BERT-WIKICREM-ALL-GAP  \\ \cite{kocijan-etal-2019-wikicrem}\end{tabular} & 76.7 & 79.4 & 1.04 & \textbf{78.0} \\
\rowcolor{LightCyan} \begin{tabular}{@{}l@{}}BERT-WIKICREM  \\ \cite{kocijan-etal-2019-wikicrem}\end{tabular} & 60.5 & 57.5 & 0.95 & 59.0 \\
\rowcolor{LightCyan} \textbf{MNPP (this work)} & 71.3 & 75.2 & 1.05 & \textbf{73.3} \\ \bottomrule
\end{tabular}
}
\caption{\small Performance comparisons among previous works and MNPP on GAP measured in F1. M stands for male, F stands for female, B stands for bias, and O stands for overall. Works highlighted with lightgray are supervised methods and works highlighted with cyan are fully un-supervised methods.}
\label{tab:gap}
% }
\end{table}

Compared with previous methods in Table \ref{tab:unsupcompare}, MNPP outperforms all unsupervised methods on all datasets and is comparable with several strong supervised methods. Current best unsupervised methods on WinoGrande is either random guess or below it, however, MNPP outperforms all of them by a margin of at least 8\%. Even compared with a supervised baseline where BERT is first finetuned on DPR, our method outperforms it by 8\%. On WSC, MNPP also outperforms all SOTA unsupervised methods by more than 8\% and outperforms most supervised methods by at least 4\% except RoBERTa-large finetuned on another pronoun resolution dataset. On DPR, our method outperforms the SOTA unsupervised baseline over 3\% and also achieves only 1\% behind the strong supervised baseline that finetunes BERT on MaskedWiki and DPR sequentially or only on WinoGrande. On KnowRef, MNPP outperforms the only unsupervised baseline by nearly 15\% and achieves only 5\% behind SOTA supervised model. Finally, on COPA, we show that MNPP gives models better common sense knowledge than finetuning on WinoGrande.

Meanwhile, we are not surprised that SOTA supervised methods still outperform unsupervised methods, including ours, considering the supervision itself and huge models with billions of parameters such as T5-11B.

% \noindent
% \textbf{Few-Shot Pronoun Resolution:}
\subsection{Few-Shot Pronoun Resolution}
\begin{figure}[t!]
\centering
\includegraphics[width=0.9\linewidth]{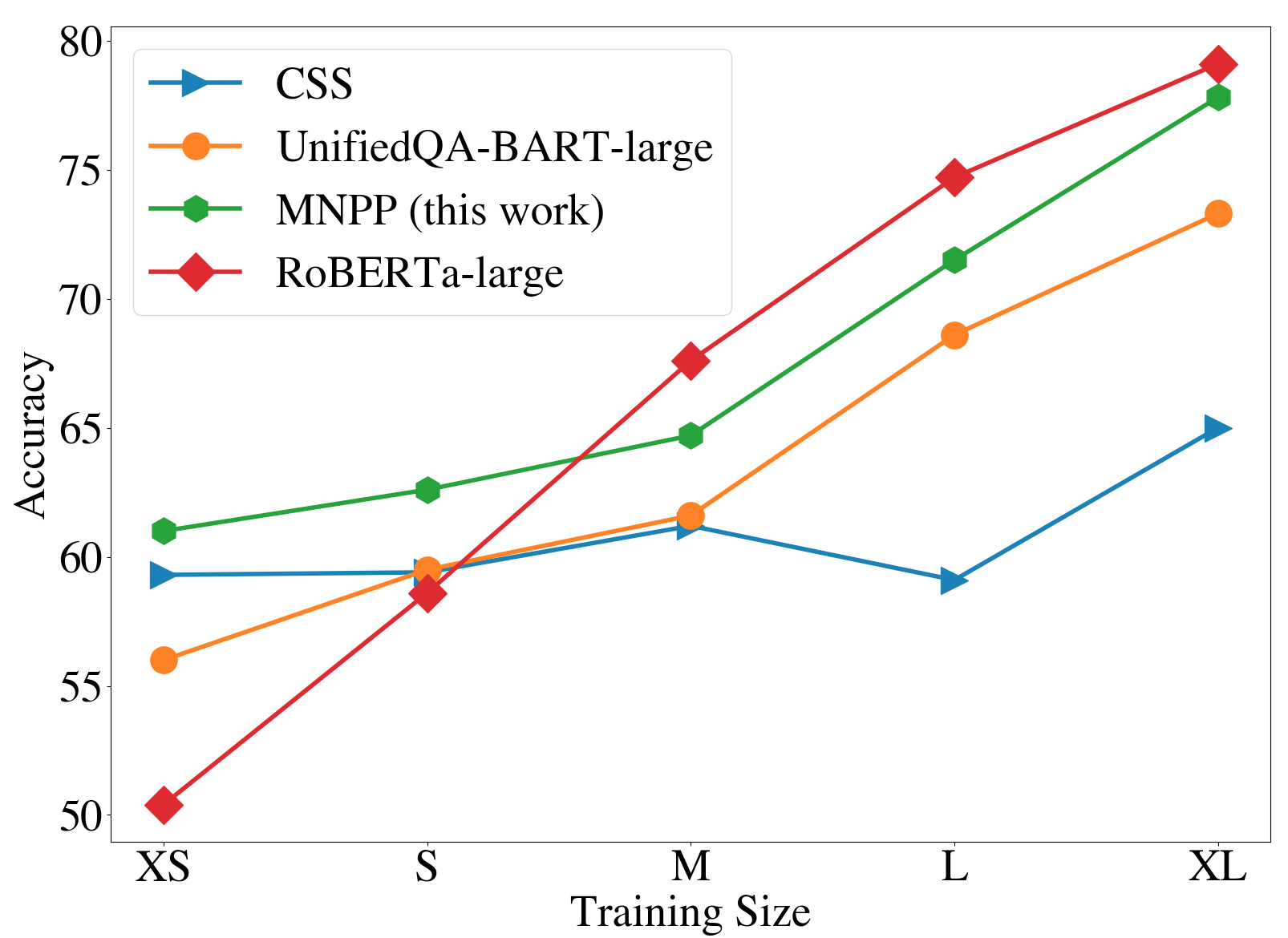}
\caption{\small Performances (\%) on WinoGrande test set after finetuning on 5 sizes of WinoGrande training set.}
\label{fig:few-shot}
\end{figure}

We further proceed to the few-shot setting on WinoGrande-S and XS. We take the top three performance zero-shot models on WinoGrande development set and finetune them on WinoGrande-XS (160 instances) and S (640 instances) separately. After few-shot evaluation, we also finetune on the remaining three data splits, which are WinoGrande-M, L, and XL. Best performances on all 5 data splits are reported in Fig. \ref{fig:few-shot} and AUC scores are reported in thrid column of 
WinoGrande section in Table \ref{tab:unsupcompare}.
% Experiment setting details are reported in Appx. \ref{ssec:fewexpdetails}. 

% \noindent
% \textbf{Results:} 
\subsubsection{Results}
As indicated in Figure \ref{fig:few-shot}, MNPP outperforms CCS, UnifiedQA-BART-large, and RoBERTa-large on WinoGrande-S and XS with a large margin, and more importantly, achieves a higher AUC score as indicated in Table \ref{tab:unsupcompare}. It is clear that MNPP pre-training gives the model crucial additional information in the few-shot setting where only minimal data is available. We also notice that in the AUC column of Table \ref{tab:unsup}, there is a negative correlation between zero-shot performance and AUC score, which means higher zero-shot performance does not guarantee better finetuning results. 

Again we need to mention that we are not comparing with SOTA performances from billions-parameters models such as UnifiedQA-T5-11B from \citet{khashabi-etal-2020-unifiedqa} or T5-3B from \citet{lin2020tttttackling}.

\section{Conclusion}
In this work, we propose MNPP pre-training to tackle unsupervised pronoun resolution and study how different properties of the synthetic pre-training dataset impact zero-shot performance on downstream datasets. Without finetuning on any pronoun resolution signal, MNPP outperforms all previous fully unsupervised methods on all tasks we study and even several strong supervised baselines. In the few-shot case where we finetune the zero-shot transfer model on WinoGrande-S and XS respectively, our model outperforms baselines by large margins, and further achieves a higher AUC score.

This work shows the effectiveness of unsupervised task definitions on text-based pronoun-resolution and common sense reasoning tasks. It would be interesting to design such tasks for multi-modal common sense reasoning~\cite{Zellers2019FromRT, fang-etal-2020-video2commonsense}.

\section*{Acknowledgements}
The authors acknowledge support from the DARPA SAIL-ON program W911NF2020006, ONR award N00014-20-1-2332, and NSF grant 1816039; and thank Yulong Chen for proofreading and the anonymous reviewers for their insightful discussion.

% \section*{Acknowledgments}

% The acknowledgments should go immediately before the references. Do not number the acknowledgments section.
% \textbf{Do not include this section when submitting your paper for review.}

\bibliographystyle{acl_natbib}
\bibliography{acl2021}

\begin{thebibliography}{48}
\expandafter\ifx\csname natexlab\endcsname\relax\def\natexlab#1{#1}\fi

\bibitem[{Banerjee and Baral(2020)}]{banerjee-baral-2020-self}
Pratyay Banerjee and Chitta Baral. 2020.
\newblock \href {https://doi.org/10.18653/v1/2020.emnlp-main.11}
  {Self-supervised knowledge triplet learning for zero-shot question
  answering}.
\newblock In \emph{Proceedings of the 2020 Conference on Empirical Methods in
  Natural Language Processing (EMNLP)}, pages 151--162, Online. Association for
  Computational Linguistics.

\bibitem[{Banerjee et~al.(2021)Banerjee, Gokhale, and
  Baral}]{banerjee-etal-2021-self}
Pratyay Banerjee, Tejas Gokhale, and Chitta Baral. 2021.
\newblock \href {https://www.aclweb.org/anthology/2021.naacl-main.95}
  {Self-supervised test-time learning for reading comprehension}.
\newblock In \emph{Proceedings of the 2021 Conference of the North American
  Chapter of the Association for Computational Linguistics: Human Language
  Technologies}, pages 1200--1211, Online. Association for Computational
  Linguistics.

\bibitem[{Banerjee et~al.(2020)Banerjee, Gokhale, Yang, and
  Baral}]{banerjee2020self}
Pratyay Banerjee, Tejas Gokhale, Yezhou Yang, and Chitta Baral. 2020.
\newblock \href {https://arxiv.org/abs/2012.02356.pdf} {Self-supervised vqa:
  Answering visual questions using images and captions}.
\newblock \emph{arXiv preprint arXiv:2012.02356}.

\bibitem[{Bowman et~al.(2015)Bowman, Angeli, Potts, and
  Manning}]{bowman-etal-2015-large}
Samuel~R. Bowman, Gabor Angeli, Christopher Potts, and Christopher~D. Manning.
  2015.
\newblock \href {https://doi.org/10.18653/v1/D15-1075} {A large annotated
  corpus for learning natural language inference}.
\newblock In \emph{Proceedings of the 2015 Conference on Empirical Methods in
  Natural Language Processing}, pages 632--642, Lisbon, Portugal. Association
  for Computational Linguistics.

\bibitem[{Dasigi et~al.(2019)Dasigi, Liu, Marasovi{\'c}, Smith, and
  Gardner}]{dasigi-etal-2019-quoref}
Pradeep Dasigi, Nelson~F. Liu, Ana Marasovi{\'c}, Noah~A. Smith, and Matt
  Gardner. 2019.
\newblock \href {https://doi.org/10.18653/v1/D19-1606} {{Q}uoref: A reading
  comprehension dataset with questions requiring coreferential reasoning}.
\newblock In \emph{Proceedings of the 2019 Conference on Empirical Methods in
  Natural Language Processing and the 9th International Joint Conference on
  Natural Language Processing (EMNLP-IJCNLP)}, pages 5925--5932, Hong Kong,
  China. Association for Computational Linguistics.

\bibitem[{Devlin et~al.(2019)Devlin, Chang, Lee, and
  Toutanova}]{devlin-etal-2019-bert}
Jacob Devlin, Ming-Wei Chang, Kenton Lee, and Kristina Toutanova. 2019.
\newblock \href {https://doi.org/10.18653/v1/N19-1423} {{BERT}: Pre-training of
  deep bidirectional transformers for language understanding}.
\newblock In \emph{Proceedings of the 2019 Conference of the North {A}merican
  Chapter of the Association for Computational Linguistics: Human Language
  Technologies, Volume 1 (Long and Short Papers)}, pages 4171--4186,
  Minneapolis, Minnesota. Association for Computational Linguistics.

\bibitem[{Emami et~al.(2019)Emami, Trichelair, Trischler, Suleman, Schulz, and
  Cheung}]{emami-etal-2019-knowref}
Ali Emami, Paul Trichelair, Adam Trischler, Kaheer Suleman, Hannes Schulz, and
  Jackie Chi~Kit Cheung. 2019.
\newblock \href {https://doi.org/10.18653/v1/P19-1386} {The {K}now{R}ef
  coreference corpus: Removing gender and number cues for difficult pronominal
  anaphora resolution}.
\newblock In \emph{Proceedings of the 57th Annual Meeting of the Association
  for Computational Linguistics}, pages 3952--3961, Florence, Italy.
  Association for Computational Linguistics.

\bibitem[{Fang et~al.(2020)Fang, Gokhale, Banerjee, Baral, and
  Yang}]{fang-etal-2020-video2commonsense}
Zhiyuan Fang, Tejas Gokhale, Pratyay Banerjee, Chitta Baral, and Yezhou Yang.
  2020.
\newblock \href {https://doi.org/10.18653/v1/2020.emnlp-main.61}
  {{V}ideo2{C}ommonsense: Generating commonsense descriptions to enrich video
  captioning}.
\newblock In \emph{Proceedings of the 2020 Conference on Empirical Methods in
  Natural Language Processing (EMNLP)}, pages 840--860, Online. Association for
  Computational Linguistics.

\bibitem[{Gururangan et~al.(2018)Gururangan, Swayamdipta, Levy, Schwartz,
  Bowman, and Smith}]{gururangan-etal-2018-annotation}
Suchin Gururangan, Swabha Swayamdipta, Omer Levy, Roy Schwartz, Samuel Bowman,
  and Noah~A. Smith. 2018.
\newblock \href {https://doi.org/10.18653/v1/N18-2017} {Annotation artifacts in
  natural language inference data}.
\newblock In \emph{Proceedings of the 2018 Conference of the North {A}merican
  Chapter of the Association for Computational Linguistics: Human Language
  Technologies, Volume 2 (Short Papers)}, pages 107--112, New Orleans,
  Louisiana. Association for Computational Linguistics.

\bibitem[{He et~al.(2019)He, Liu, Chen, and Gao}]{he2019hybrid}
Pengcheng He, Xiaodong Liu, Weizhu Chen, and Jianfeng Gao. 2019.
\newblock \href {https://arxiv.org/pdf/1907.11983.pdf} {A hybrid neural network
  model for commonsense reasoning}.
\newblock \emph{arXiv preprint arXiv:1907.11983}.

\bibitem[{Hochreiter and Schmidhuber(1997)}]{hochreiter1997long}
Sepp Hochreiter and J{\"u}rgen Schmidhuber. 1997.
\newblock \href
  {http://citeseerx.ist.psu.edu/viewdoc/download?doi=10.1.1.676.4320\&rep=rep1\&type=pdf}
  {Long short-term memory}.
\newblock \emph{Neural computation}, 9(8):1735--1780.

\bibitem[{Khashabi et~al.(2020)Khashabi, Min, Khot, Sabharwal, Tafjord, Clark,
  and Hajishirzi}]{khashabi-etal-2020-unifiedqa}
Daniel Khashabi, Sewon Min, Tushar Khot, Ashish Sabharwal, Oyvind Tafjord,
  Peter Clark, and Hannaneh Hajishirzi. 2020.
\newblock \href {https://doi.org/10.18653/v1/2020.findings-emnlp.171}
  {{UNIFIEDQA}: Crossing format boundaries with a single {QA} system}.
\newblock In \emph{Findings of the Association for Computational Linguistics:
  EMNLP 2020}, pages 1896--1907, Online. Association for Computational
  Linguistics.

\bibitem[{Khot et~al.(2020)Khot, Clark, Guerquin, Jansen, and
  Sabharwal}]{khot2020qasc}
Tushar Khot, Peter Clark, Michal Guerquin, Peter Jansen, and Ashish Sabharwal.
  2020.
\newblock \href {https://ojs.aaai.org/index.php/AAAI/article/view/6319} {Qasc:
  A dataset for question answering via sentence composition}.
\newblock In \emph{Proceedings of the AAAI Conference on Artificial
  Intelligence}, volume~34, pages 8082--8090.

\bibitem[{Klein and Nabi(2019)}]{klein-nabi-2019-attention}
Tassilo Klein and Moin Nabi. 2019.
\newblock \href {https://doi.org/10.18653/v1/P19-1477} {Attention is (not) all
  you need for commonsense reasoning}.
\newblock In \emph{Proceedings of the 57th Annual Meeting of the Association
  for Computational Linguistics}, pages 4831--4836, Florence, Italy.
  Association for Computational Linguistics.

\bibitem[{Klein and Nabi(2020)}]{klein-nabi-2020-contrastive}
Tassilo Klein and Moin Nabi. 2020.
\newblock \href {https://doi.org/10.18653/v1/2020.acl-main.671} {Contrastive
  self-supervised learning for commonsense reasoning}.
\newblock In \emph{Proceedings of the 58th Annual Meeting of the Association
  for Computational Linguistics}, pages 7517--7523, Online. Association for
  Computational Linguistics.

\bibitem[{Kocijan et~al.(2019{\natexlab{a}})Kocijan, Camburu, Cretu, Yordanov,
  Blunsom, and Lukasiewicz}]{kocijan-etal-2019-wikicrem}
Vid Kocijan, Oana-Maria Camburu, Ana-Maria Cretu, Yordan Yordanov, Phil
  Blunsom, and Thomas Lukasiewicz. 2019{\natexlab{a}}.
\newblock \href {https://doi.org/10.18653/v1/D19-1439} {{W}iki{CREM}: A large
  unsupervised corpus for coreference resolution}.
\newblock In \emph{Proceedings of the 2019 Conference on Empirical Methods in
  Natural Language Processing and the 9th International Joint Conference on
  Natural Language Processing (EMNLP-IJCNLP)}, pages 4303--4312, Hong Kong,
  China. Association for Computational Linguistics.

\bibitem[{Kocijan et~al.(2019{\natexlab{b}})Kocijan, Cretu, Camburu, Yordanov,
  and Lukasiewicz}]{kocijan-etal-2019-surprisingly}
Vid Kocijan, Ana-Maria Cretu, Oana-Maria Camburu, Yordan Yordanov, and Thomas
  Lukasiewicz. 2019{\natexlab{b}}.
\newblock \href {https://doi.org/10.18653/v1/P19-1478} {A surprisingly robust
  trick for the {W}inograd schema challenge}.
\newblock In \emph{Proceedings of the 57th Annual Meeting of the Association
  for Computational Linguistics}, pages 4837--4842, Florence, Italy.
  Association for Computational Linguistics.

\bibitem[{Kocijan et~al.(2020)Kocijan, Lukasiewicz, Davis, Marcus, and
  Morgenstern}]{kocijan2020review}
Vid Kocijan, Thomas Lukasiewicz, Ernest Davis, Gary Marcus, and Leora
  Morgenstern. 2020.
\newblock \href {https://arxiv.org/pdf/2004.13831.pdf} {A review of winograd
  schema challenge datasets and approaches}.
\newblock \emph{arXiv preprint arXiv:2004.13831}.

\bibitem[{Lai et~al.(2017)Lai, Xie, Liu, Yang, and Hovy}]{lai-etal-2017-race}
Guokun Lai, Qizhe Xie, Hanxiao Liu, Yiming Yang, and Eduard Hovy. 2017.
\newblock \href {https://doi.org/10.18653/v1/D17-1082} {{RACE}: Large-scale
  {R}e{A}ding comprehension dataset from examinations}.
\newblock In \emph{Proceedings of the 2017 Conference on Empirical Methods in
  Natural Language Processing}, pages 785--794, Copenhagen, Denmark.
  Association for Computational Linguistics.

\bibitem[{Lee et~al.(2018)Lee, He, and Zettlemoyer}]{lee-etal-2018-higher}
Kenton Lee, Luheng He, and Luke Zettlemoyer. 2018.
\newblock \href {https://doi.org/10.18653/v1/N18-2108} {Higher-order
  coreference resolution with coarse-to-fine inference}.
\newblock In \emph{Proceedings of the 2018 Conference of the North {A}merican
  Chapter of the Association for Computational Linguistics: Human Language
  Technologies, Volume 2 (Short Papers)}, pages 687--692, New Orleans,
  Louisiana. Association for Computational Linguistics.

\bibitem[{Levesque et~al.(2012)Levesque, Davis, and
  Morgenstern}]{levesque2012winograd}
Hector Levesque, Ernest Davis, and Leora Morgenstern. 2012.
\newblock \href
  {https://citeseerx.ist.psu.edu/viewdoc/download?doi=10.1.1.729.9814&rep=rep1&type=pdf}
  {The winograd schema challenge}.
\newblock In \emph{Thirteenth International Conference on the Principles of
  Knowledge Representation and Reasoning}. Citeseer.

\bibitem[{Lewis et~al.(2019)Lewis, Denoyer, and
  Riedel}]{lewis-etal-2019-unsupervised}
Patrick Lewis, Ludovic Denoyer, and Sebastian Riedel. 2019.
\newblock \href {https://doi.org/10.18653/v1/P19-1484} {Unsupervised question
  answering by cloze translation}.
\newblock In \emph{Proceedings of the 57th Annual Meeting of the Association
  for Computational Linguistics}, pages 4896--4910, Florence, Italy.
  Association for Computational Linguistics.

\bibitem[{Li et~al.(2020)Li, Wang, Dong, Wei, and Xu}]{li-etal-2020-harvesting}
Zhongli Li, Wenhui Wang, Li~Dong, Furu Wei, and Ke~Xu. 2020.
\newblock \href {https://doi.org/10.18653/v1/2020.acl-main.600} {Harvesting and
  refining question-answer pairs for unsupervised {QA}}.
\newblock In \emph{Proceedings of the 58th Annual Meeting of the Association
  for Computational Linguistics}, pages 6719--6728, Online. Association for
  Computational Linguistics.

\bibitem[{Lin et~al.(2020)Lin, Yang, Nogueira, Tsai, Wang, and
  Lin}]{lin2020tttttackling}
Sheng-Chieh Lin, Jheng-Hong Yang, Rodrigo Nogueira, Ming-Feng Tsai, Chuan-Ju
  Wang, and Jimmy Lin. 2020.
\newblock \href {https://arxiv.org/pdf/2003.08380.pdf} {Tttttackling winogrande
  schemas}.
\newblock \emph{arXiv preprint arXiv:2003.08380}.

\bibitem[{Liu et~al.(2019)Liu, Ott, Goyal, Du, Joshi, Chen, Levy, Lewis,
  Zettlemoyer, and Stoyanov}]{liu2019roberta}
Yinhan Liu, Myle Ott, Naman Goyal, Jingfei Du, Mandar Joshi, Danqi Chen, Omer
  Levy, Mike Lewis, Luke Zettlemoyer, and Veselin Stoyanov. 2019.
\newblock \href {https://arxiv.org/pdf/1907.11692.pdf} {Roberta: A robustly
  optimized bert pretraining approach}.
\newblock \emph{arXiv preprint arXiv:1907.11692}.

\bibitem[{Mikolov et~al.(2013)Mikolov, Chen, Corrado, and
  Dean}]{mikolov2013efficient}
Tomas Mikolov, Kai Chen, Greg Corrado, and Jeffrey Dean. 2013.
\newblock \href {https://arxiv.org/abs/1301.3781} {Efficient estimation of word
  representations in vector space}.
\newblock \emph{arXiv preprint arXiv:1301.3781}.

\bibitem[{Morgenstern et~al.(2016)Morgenstern, Davis, and
  Ortiz}]{morgenstern2016planning}
Leora Morgenstern, Ernest Davis, and Charles~L Ortiz. 2016.
\newblock \href {https://ojs.aaai.org/index.php/aimagazine/article/view/2639}
  {Planning, executing, and evaluating the winograd schema challenge}.
\newblock \emph{AI Magazine}, 37(1):50--54.

\bibitem[{Opitz and Frank(2018)}]{opitz-frank-2018-addressing}
Juri Opitz and Anette Frank. 2018.
\newblock \href {https://www.aclweb.org/anthology/W18-4105} {Addressing the
  {W}inograd schema challenge as a sequence ranking task}.
\newblock In \emph{Proceedings of the First International Workshop on Language
  Cognition and Computational Models}, pages 41--52, Santa Fe, New Mexico, USA.
  Association for Computational Linguistics.

\bibitem[{Pradhan et~al.(2012)Pradhan, Moschitti, Xue, Uryupina, and
  Zhang}]{pradhan-etal-2012-conll}
Sameer Pradhan, Alessandro Moschitti, Nianwen Xue, Olga Uryupina, and Yuchen
  Zhang. 2012.
\newblock \href {https://www.aclweb.org/anthology/W12-4501} {{C}o{NLL}-2012
  shared task: Modeling multilingual unrestricted coreference in
  {O}nto{N}otes}.
\newblock In \emph{Joint Conference on {EMNLP} and {C}o{NLL} - Shared Task},
  pages 1--40, Jeju Island, Korea. Association for Computational Linguistics.

\bibitem[{Prakash et~al.(2019)Prakash, Sharma, Mitra, and
  Baral}]{prakash-etal-2019-combining}
Ashok Prakash, Arpit Sharma, Arindam Mitra, and Chitta Baral. 2019.
\newblock \href {https://doi.org/10.18653/v1/P19-1614} {Combining knowledge
  hunting and neural language models to solve the {W}inograd schema challenge}.
\newblock In \emph{Proceedings of the 57th Annual Meeting of the Association
  for Computational Linguistics}, pages 6110--6119, Florence, Italy.
  Association for Computational Linguistics.

\bibitem[{Radford et~al.(2019)Radford, Wu, Child, Luan, Amodei, and
  Sutskever}]{radford2019language}
Alec Radford, Jeffrey Wu, Rewon Child, David Luan, Dario Amodei, and Ilya
  Sutskever. 2019.
\newblock \href {http://www.persagen.com/files/misc/radford2019language.pdf}
  {Language models are unsupervised multitask learners}.
\newblock \emph{OpenAI blog}, 1(8):9.

\bibitem[{Rae et~al.(2020)Rae, Potapenko, Jayakumar, Hillier, and
  Lillicrap}]{Rae2020Compressive}
Jack~W. Rae, Anna Potapenko, Siddhant~M. Jayakumar, Chloe Hillier, and
  Timothy~P. Lillicrap. 2020.
\newblock \href {https://openreview.net/forum?id=SylKikSYDH} {Compressive
  transformers for long-range sequence modelling}.
\newblock In \emph{International Conference on Learning Representations}.

\bibitem[{Rahman and Ng(2012)}]{rahman-ng-2012-resolving}
Altaf Rahman and Vincent Ng. 2012.
\newblock \href {https://www.aclweb.org/anthology/D12-1071} {Resolving complex
  cases of definite pronouns: The {W}inograd schema challenge}.
\newblock In \emph{Proceedings of the 2012 Joint Conference on Empirical
  Methods in Natural Language Processing and Computational Natural Language
  Learning}, pages 777--789, Jeju Island, Korea. Association for Computational
  Linguistics.

\bibitem[{Rajpurkar et~al.(2016)Rajpurkar, Zhang, Lopyrev, and
  Liang}]{rajpurkar-etal-2016-squad}
Pranav Rajpurkar, Jian Zhang, Konstantin Lopyrev, and Percy Liang. 2016.
\newblock \href {https://doi.org/10.18653/v1/D16-1264} {{SQ}u{AD}: 100,000+
  questions for machine comprehension of text}.
\newblock In \emph{Proceedings of the 2016 Conference on Empirical Methods in
  Natural Language Processing}, pages 2383--2392, Austin, Texas. Association
  for Computational Linguistics.

\bibitem[{Ruan et~al.(2019)Ruan, Zhu, Ling, Shi, Liu, and
  Wei}]{ruan2019exploring}
Yu-Ping Ruan, Xiaodan Zhu, Zhen-Hua Ling, Zhan Shi, Quan Liu, and Si~Wei. 2019.
\newblock \href {https://arxiv.org/pdf/1904.09705.pdf} {Exploring unsupervised
  pretraining and sentence structure modelling for winograd schema challenge}.
\newblock \emph{arXiv preprint arXiv:1904.09705}.

\bibitem[{Sakaguchi et~al.(2020)Sakaguchi, Le~Bras, Bhagavatula, and
  Choi}]{sakaguchi2020winogrande}
Keisuke Sakaguchi, Ronan Le~Bras, Chandra Bhagavatula, and Yejin Choi. 2020.
\newblock \href {https://ojs.aaai.org/index.php/AAAI/article/view/6399}
  {Winogrande: An adversarial winograd schema challenge at scale}.
\newblock In \emph{Proceedings of the AAAI Conference on Artificial
  Intelligence}, volume~34, pages 8732--8740.

\bibitem[{See et~al.(2017)See, Liu, and Manning}]{see-etal-2017-get}
Abigail See, Peter~J. Liu, and Christopher~D. Manning. 2017.
\newblock \href {https://doi.org/10.18653/v1/P17-1099} {Get to the point:
  Summarization with pointer-generator networks}.
\newblock In \emph{Proceedings of the 55th Annual Meeting of the Association
  for Computational Linguistics (Volume 1: Long Papers)}, pages 1073--1083,
  Vancouver, Canada. Association for Computational Linguistics.

\bibitem[{Speer and Havasi(2012)}]{speer2012representing}
Robert Speer and Catherine Havasi. 2012.
\newblock \href {http://lrec-conf.org/proceedings/lrec2012/pdf/1072_Paper.pdf}
  {Representing general relational knowledge in conceptnet 5.}
\newblock In \emph{LREC}, pages 3679--3686.

\bibitem[{Trinh and Le(2018)}]{trinh2018simple}
Trieu~H Trinh and Quoc~V Le. 2018.
\newblock \href {https://arxiv.org/pdf/1806.02847.pdf} {A simple method for
  commonsense reasoning}.
\newblock \emph{arXiv preprint arXiv:1806.02847}.

\bibitem[{Wang et~al.(2019{\natexlab{a}})Wang, Pruksachatkun, Nangia, Singh,
  Michael, Hill, Levy, and Bowman}]{wang2019superglue}
Alex Wang, Yada Pruksachatkun, Nikita Nangia, Amanpreet Singh, Julian Michael,
  Felix Hill, Omer Levy, and Samuel~R Bowman. 2019{\natexlab{a}}.
\newblock \href {https://arxiv.org/pdf/1905.00537.pdf} {Superglue: A stickier
  benchmark for general-purpose language understanding systems}.
\newblock \emph{arXiv preprint arXiv:1905.00537}.

\bibitem[{Wang et~al.(2019{\natexlab{b}})Wang, Zhang, Shen, Liu, Liu, Gao, and
  Jiang}]{wang-etal-2019-unsupervised}
Shuohang Wang, Sheng Zhang, Yelong Shen, Xiaodong Liu, Jingjing Liu, Jianfeng
  Gao, and Jing Jiang. 2019{\natexlab{b}}.
\newblock \href {https://doi.org/10.18653/v1/N19-1094} {Unsupervised deep
  structured semantic models for commonsense reasoning}.
\newblock In \emph{Proceedings of the 2019 Conference of the North {A}merican
  Chapter of the Association for Computational Linguistics: Human Language
  Technologies, Volume 1 (Long and Short Papers)}, pages 882--891, Minneapolis,
  Minnesota. Association for Computational Linguistics.

\bibitem[{Webster et~al.(2018)Webster, Recasens, Axelrod, and
  Baldridge}]{webster-etal-2018-mind}
Kellie Webster, Marta Recasens, Vera Axelrod, and Jason Baldridge. 2018.
\newblock \href {https://doi.org/10.1162/tacl_a_00240} {Mind the {GAP}: A
  balanced corpus of gendered ambiguous pronouns}.
\newblock \emph{Transactions of the Association for Computational Linguistics},
  6:605--617.

\bibitem[{Williams et~al.(2018)Williams, Nangia, and
  Bowman}]{williams-etal-2018-broad}
Adina Williams, Nikita Nangia, and Samuel Bowman. 2018.
\newblock \href {https://doi.org/10.18653/v1/N18-1101} {A broad-coverage
  challenge corpus for sentence understanding through inference}.
\newblock In \emph{Proceedings of the 2018 Conference of the North {A}merican
  Chapter of the Association for Computational Linguistics: Human Language
  Technologies, Volume 1 (Long Papers)}, pages 1112--1122, New Orleans,
  Louisiana. Association for Computational Linguistics.

\bibitem[{Ye et~al.(2020)Ye, Lin, Du, Liu, Li, Sun, and
  Liu}]{ye-etal-2020-coreferential}
Deming Ye, Yankai Lin, Jiaju Du, Zhenghao Liu, Peng Li, Maosong Sun, and
  Zhiyuan Liu. 2020.
\newblock \href {https://doi.org/10.18653/v1/2020.emnlp-main.582}
  {{C}oreferential {R}easoning {L}earning for {L}anguage {R}epresentation}.
\newblock In \emph{Proceedings of the 2020 Conference on Empirical Methods in
  Natural Language Processing (EMNLP)}, pages 7170--7186, Online. Association
  for Computational Linguistics.

\bibitem[{Ye et~al.(2019)Ye, Chen, Wang, and Ling}]{ye2019align}
Zhi-Xiu Ye, Qian Chen, Wen Wang, and Zhen-Hua Ling. 2019.
\newblock \href {https://arxiv.org/pdf/1908.06725.pdf} {Align, mask and select:
  A simple method for incorporating commonsense knowledge into language
  representation models}.
\newblock \emph{arXiv preprint arXiv:1908.06725}.

\bibitem[{Zellers et~al.(2019)Zellers, Bisk, Farhadi, and
  Choi}]{Zellers2019FromRT}
Rowan Zellers, Yonatan Bisk, Ali Farhadi, and Yejin Choi. 2019.
\newblock \href {https://ieeexplore.ieee.org/document/8953217} {From
  recognition to cognition: Visual commonsense reasoning}.
\newblock \emph{2019 IEEE/CVF Conference on Computer Vision and Pattern
  Recognition (CVPR)}, pages 6713--6724.

\bibitem[{Zhang and Song(2018)}]{zhang2018distributed}
Hongming Zhang and Yangqiu Song. 2018.
\newblock \href {https://dl.acm.org/doi/abs/10.1145/3195106.3195127} {A
  distributed solution for winograd schema challenge}.
\newblock In \emph{Proceedings of the 2018 10th International Conference on
  Machine Learning and Computing}, pages 322--326.

\bibitem[{Zhang et~al.(2017)Zhang, Zhong, Chen, Angeli, and
  Manning}]{zhang-etal-2017-position}
Yuhao Zhang, Victor Zhong, Danqi Chen, Gabor Angeli, and Christopher~D.
  Manning. 2017.
\newblock \href {https://doi.org/10.18653/v1/D17-1004} {Position-aware
  attention and supervised data improve slot filling}.
\newblock In \emph{Proceedings of the 2017 Conference on Empirical Methods in
  Natural Language Processing}, pages 35--45, Copenhagen, Denmark. Association
  for Computational Linguistics.

\end{thebibliography}

\clearpage

\appendix

\section{Related Work on Supervised Methods}
\label{ssec:appxrelatedwork}
\textbf{WSC \& DPR.} \citet{opitz-frank-2018-addressing} is the first work to propose transfer learning from another pronoun resolution dataset such as DPR to WSC. \citet{he2019hybrid} use a hybrid model of \citet{wang-etal-2019-unsupervised} and \citet{kocijan-etal-2019-surprisingly}. \citet{ruan2019exploring} explore BERT's next sentence prediction with finetuning on DPR. \citet{ye-etal-2020-coreferential} finetune a new language representation model called CorefBERT, which is trained with a novel task to strengthen the co-referential reasoning ability of BERT, on DPR and then test on DPR and WSC. The SOTA supervised performance is provided by \citet{sakaguchi2020winogrande} where they finetune a RoBERTa-large model on WinoGrande or DPR and evaluate on WSC and DPR without and with further finetuning. A detailed review of WSC and WSC related dataset can be found at \citet{kocijan2020review}.

\smallskip
\noindent
\textbf{KnowRef.} In \citet{emami-etal-2019-knowref}, an end-to-end neural system \cite{lee-etal-2018-higher} is trained on CoNLL2012 shared task \cite{pradhan-etal-2012-conll} and then tested under three settings: directly applying to KnowRef test set, retraining on KnowRef, and retraining on KnowRef plus CoNLL2012. \citet{sakaguchi2020winogrande} transfer a WinoGrande-finetuned RoBERTa-large model and DPR-finetuned RoBERTa-large model to KnowRef test set respectively.

\smallskip
\noindent
\textbf{WinoGrande.} The authors of WinoGrande finetune a RoBERTa-large on WinoGrande training set and evaluate on the test set in standard supervised setting, and \citet{lin2020tttttackling} finetune a T5-3B model instead. \citet{sakaguchi2020winogrande} also study finetuning BERT and RoBERTa with only local context (only tokens near the pronoun location are available instead of the whole sentence). \citet{ye-etal-2020-coreferential} finetune WinoGrande using CorefBERT. \citet{klein-nabi-2020-contrastive} finetune their unsupervised CSS model. Finally, UnifiedQA \cite{khashabi-etal-2020-unifiedqa}, which is pre-trained on eight seed QA datasets spanning four different formats in a unified way, is finetuned on WinoGrande.

\section{Synthetic Datasets Construction}
\label{ssec:appxsynthdataset}
For the first synthetic dataset in the first group, we choose 5000 stories in CNN stories, a small portion of Gutenberg books, and the whole training set of QUOREF \cite{dasigi-etal-2019-quoref}, which is a reading comprehension dataset that requires resolving co-reference among entities crawled from Wikipedia, and these sources result in the size of 160k. The second synthetic dataset in the first group comprises the same sources as above plus extra knowledge crawled by Google query using the knowledge hunting strategy introduced in \citet{prakash-etal-2019-combining}. Following their strategy, we scrap 6531 and 69462 knowledge sentences for WSC and WinoGrande respectively. We relax the filtering process to allow longer sentences than those in the first synthetic dataset and lead to 380k samples in total. We then fix the text style and study the influence of data size on pre-training. We use 2000 books from PG-19 as the source and create five synthetic datasets with size of 500k, 300k, 100k, 50k, and 10k as the second group. We further study how difficulty levels of samples affect the downstream zero-shot performance. We select 100k samples from the PG-19 books described above and evenly split them into three synthetic datasets with low, medium, and high similarity scores between candidate choices as the third group. As a result, we create 3 groups of synthetic datasets with ten synthetic datasets in total. We used spaCy\footnote{\small{\url{https://spacy.io/}}} to pre-process raw text, including removing blank spaces, special characters, sentences that are too short or too long, and extracting noun-phrases.

\section{Zero-shot Experiment Details}
\label{ssec:zeroexpdetails}
Recent study \cite{khot2020qasc} has shown that finetuning a RACE-finetuned \cite{lai-etal-2017-race} RoBERTa model as a start point is much more stable than directly finetuning a RoBERTa model from scratch, we follow the same strategy to start finetuning a RACE-finetuned RoBERTa-large model on all synthetic datasets. We use Hugging Face Transformers\footnote{\small{\url{https://github.com/huggingface/}}} as our codebase. We set Adam optimizer with an initial learning rate of $1e-5$ and epsilon of $1e-8$, and without weight decaying for all settings. For a synthetic dataset whose size is larger or equal to 100k, we choose the batch size of 32 and train for 20 epochs, otherwise, we choose the batch size of 16 and train for 50 epochs. We checkpoint every X steps, with X in [50,500].

\section{Few-shot Experiment Details}
\label{ssec:fewexpdetails}
We set Adam optimizer with an initial learning rate of $1e-5$ and epsilon of $1e-8$, without weight decaying, and batch size between 16 and 32 for all sizes. We finetune 20 epochs for WinoGrande-XL, L, and M, 40 epochs for S, and 160 epochs for XS. We checkpoint every X steps, with X in [50,500].

\end{document}